\pdfoutput=1

\documentclass[11pt]{article}

\usepackage[final]{acl}

\usepackage{times}
\usepackage{latexsym}
\usepackage{multirow}

\usepackage[T1]{fontenc}

\usepackage[utf8]{inputenc}

\usepackage{microtype}

\usepackage{inconsolata}

\usepackage{graphicx}
\usepackage{tablefootnote}
\usepackage{enumitem}

\title{`Finance Wizard' at the FinLLM Challenge Task: Financial Text Summarization}

\author{Meisin Lee \and Soon Lay-Ki\\
  Monash University, Bandar Sunway, Selangor Malaysia \\
 \texttt{\{lee.meisin, soon.layki\}@monash.edu}}

\begin{document}
\maketitle
\begin{abstract}
This paper presents our participation under the team name `Finance Wizard' in the FinNLP-AgentScen\footnote{\url{https://sites.google.com/nlg.csie.ntu.edu.tw/finnlp-agentscen/shared-task-finllm}} 2024 shared task \#2: Financial Text Summarization. It documents our pipeline approach of fine-tuning a foundation model into a task-specific model for Financial Text Summarization. It involves (1) adapting Llama3 8B, a foundation model, to the Finance domain via continued pre-training, (2) multi-task instruction-tuning to further equip the model with more finance-related capabilities, (3) finally fine-tuning the model into a task-specific `expert'. Our model, FinLlama3\_sum, yielded commendable results, securing the third position in its category with a ROUGE-1 score of 0.521.

\end{abstract}

\section{Introduction}
Since the release of Large Language Models (LLMs), they have been swiftly fine-tuned into specialized LLMs in various domains such as biomedical, legal, finance and economics. For the Finance domain, a number of FinLLMs have been released: \textbf{FinMA} (also known as PIXIU) \citep{xie2023pixiu}, \textbf{InvestLM} \citep{yang2023investlm}, \textbf{FinGPT} \citep{wang2023fingpt} and \textbf{BloombergGPT} \citep{wu2023bloomberggpt}. As part of expanding the capabilities of Financial LLM, organizers of the FinLLM challenge prepared three subtasks, namely financial classification, financial text summarization and stock trading prediction. For subtask 2 (financial text summarization), participants are given a dataset of financial news article and the task is to produce abstractive summaries for each piece of news.

While all the FinLLMs listed above are trained to be multi-task models capable of a wide variety of finance-related tasks, this paper differs from these FinLLMs where we aim to produce a task-specific model to achieve the best possible score for Financial Text Summarization. The training approach we took is influenced by the findings in \citep{jang2023exploring, lee2024survey} that \textit{expert} Language Models (LM) fine-tuned on just a single task can outperform a multi-task LM trained with numerous different tasks. We aim to produce a task-specific model, similar to FinPythia\citep{xie2023efficient} - a model trained specifically for financial sentiment analysis.

This paper documents the considerations and planning that went into producing the task-specific model for the financial text summarization task. This includes the selection of the Foundation Model (both evaluating existing FinLLMs and new LLMs), the selection of training corpus and datasets, as well as the design of the end-to-end fine-tuning approach.


\section{Related Work}
To the best of our knowledge, there are two survey papers written on Financial Large Language Models (FinLLMs): (1) \textit{Large Language Models in Finance: A Survey}\citep{li2023large} and (2) \textit{A survey of Large Language Models in Finance (FinLLMs)} \citep{lee2024survey}. Based on their findings, the list of FinLLMs and their properties are captured in Table \ref{table:FinLLM}. For brevity, only LLMs\footnote{According to \citep{zhao2023survey}, Large LM are models that have 7 billion or more parameters} are listed, finance PLMs (Pretrained Language Models) such as FinBERT\citep{araci2019finbert, yang2020finbert, liu2021finbert} and FLANG\citep{chung2024scaling} are omitted.  

\begin{table*}
\small
\centering
  \begin{tabular}{lllllll}
    \hline
    \textbf{Type} & 
    \textbf{Model} & \textbf{Foundation} & \textbf{Parameters} & \textbf{Techniques} & \textbf{Task(s)} & \textbf{Context Len.}\\
    \hline
    \multirow{9}{*}{\rotatebox[origin=c]{90}{Multi-task}}  & BloombergGPT & BLOOM & 50B & PT, PE & SA, TC & 2048  \\   
    & \citep{wu2023bloomberggpt} &  &  & & NER,QA &   \\
    \cline{2-7}
    & InvestLM & LLaMa & 65B & IFT, PE, & SA,TC & 2048  \\
    & \citep{yang2023investlm} & & & & QA,Summ &   \\
    \cline{2-7}
    & FinMA & LLaMa2 & 7B, 30B & IFT, PE & SA,TC,NER & 2048  \\
    & \citep{xie2023pixiu} & & & & QA,SMP &   \\
    \cline{2-7}
    & FinGPT & LlaMa2, Falcon,& 7B and & IFT, PE& SA,TC & 4096  \\
    & \citep{wang2023fingpt} & MPT, BLOOM & various sizes & & NER,RE&  \\
    &  & ChatGLM2, QWEN  &  & & &  \\
    \hline
    \multirow{2}{*}{\rotatebox[origin=c]{90}{1-Task}}  & Fin-Pythia-1.4B & Pythia & 1.4B & PT, PE & SA & 2048  \\    
    & \citep{rodriguez-inserte-etal-2023-large} &  &  & & &   \\
    \hline
  \end{tabular}
  \caption{\label{table:FinLLM} A summary of  the list of Instruction Fine-tuned FinLLMs. List of tasks that these models are fine-tuned on are: SA (Sentiment Analysis), TC (Text Classification), SBD (Structure Boundary Detection), NER (Named Entity Recognition), QA (Question Answering), SMP (Stock Movement Prediction), Summ (Text Summarization), and RE (Relation Extraction).}
  \vspace{-0.5em}
\end{table*}

These FinLLMs are trained on multiple tasks, which includes Sentiment Analysis, Text Classification, Structure Boundary Detection, Named Entity Recognition, Question Answering and more. Just like the foundation LLMs \citep{radford2019language}, these multi-task FinLLMs are also used in in-context learning and in unseen tasks via few-shot and zero-shot prompting.

\section{Approach} \label{sec:approach}
\begin{figure*}[htp]
    \centering
    \fbox{\includegraphics[width=\textwidth]{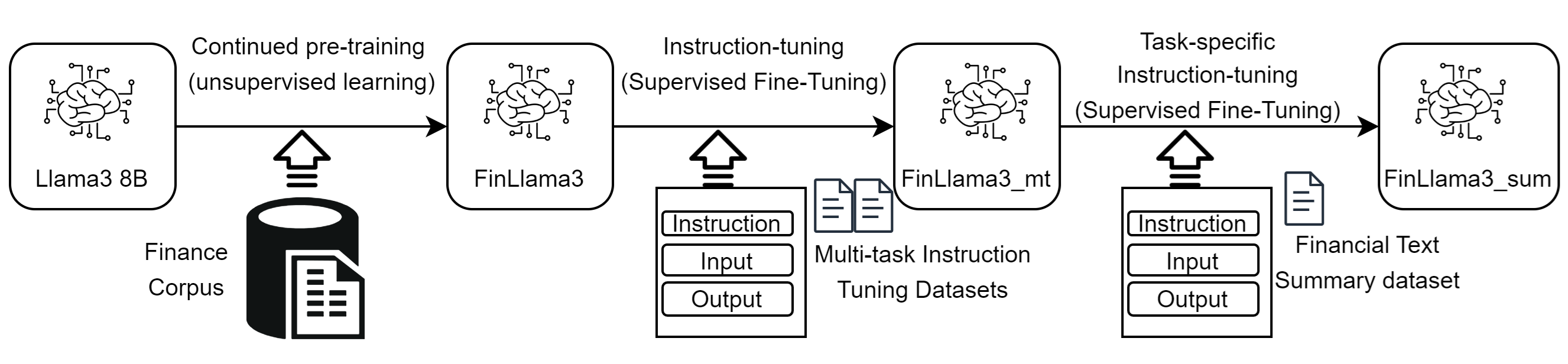}}
    \caption{The design of our end-to-end fine-tuning approach. This shows the evolution of a foundation model to the final task-specific `expert' for financial text summarization.}
    \label{fig:pipeline}
    \vspace{-0.5em}
    \end{figure*}

We designed a pipeline approach where it first starts with a foundation LLM, followed by adapting it to the finance domain, and then instruction-tuning it into a task-specific ‘expert'. This approach is illustrated in Figure \ref{fig:pipeline}, and also described below:
\begin{enumerate}[noitemsep,nolistsep]
    \item Continual pre-training: in the first stage, we further pre-trained a foundation LLM on a collection of Finance-related corpus on the text completion task (causal language modeling) via unsupervised training. This produced \textbf{FinLlama3}.
    \item Multi-task instruction-tuning: the second stage involves instruction-tuning FinLlama3 on multiple financial tasks via supervised fine-tuning (SFT) to obtain \textbf{FinLlama3\_mt}.
    \item Task-specific Instruction-tuning: in the last stage, the model is instruction-tuned specifically on just one-task - text summarization to produce \textbf{FinLlama3\_sum}.
\end{enumerate}

\subsection{Choice of Foundation Model}
For the task of Financial Text Summarization, an LLM's contextual window length plays an important role because the context window need to be large enough to fit the entire news passage so that crucial information is not lost.
The open-source Multi-task FinLLMs (FinMA, InvestLM, FinGPT) listed in Table \ref{table:FinLLM} act as good candidates as `foundation model' for instruction-tuning for Financial Text Summarization. However, their context window length is constrained to be at a maximum of 4,096 (see column `Context Len' in Table \ref{table:FinLLM} for information). At the point of writing, Llama3 (8B, 70B) is the newest open-source LLM that has the largest context window (8,192 tokens), it is boasted to outperform all of its predecessors. Hence, we chose Llama3 8B as the foundation model.

\subsection{Continual Pre-training}
Continual pre-training of LLMs aims to adapt an existing general LLM, which is also known as Foundation Models, to a new domain and be enriched with new domain knowledge, in the same way that FinBERT was trained. With Llama3 8B identified as the foundation model, we further pre-trained it on a Finance Corpus we have identified. The corpus is made up of financial texts such as financial news, financial statements, company annual reports, financial research reports, financial literature, market data, etc. Specifically, the corpus is made up of:
\begin{enumerate}[noitemsep,nolistsep]
    \item News from Reuters\footnote{\url{https://huggingface.co/datasets/rjjan/reuters21578}}:  The datasets contains 55,700 records of financial news stories from Reuters.    
    \item News from CNBC, Reuters, WSJ\footnote{\url{https://huggingface.co/datasets/Lettria/financial-articles}}:  This dataset contains 18,400 records of financial news from various sources,. 
    \item Finance and economic glossary (Investopedia)\footnote{\url{https://huggingface.co/datasets/infCapital/investopedia_terms_en}}: This dataset contains 6,310 records of financial and economic glossary 
    \item Edgar SEC filings (1993 - 2020)\footnote{\url{https://huggingface.co/datasets/eloukas/edgar-corpus}}: This dataset comprises financial reports (10-K) submitted to the U.S. Securities and Exchange Commission (SEC). We only used reports for the year 2019. 

    \item FinWeb\footnote{\url{https://github.com/deep-over/FiLM/tree/main/pretraining/document/train_dataset/finweb}}: The dataset is a website that provides economic knowledge and information on finance, loans and products.    
\end{enumerate}

\subsection{Instruction Fine-tuning}
Instruction Fine-tuning fine-tunes LLMs on a stream of supervised instruction-following data, this step aims to empower LLMs to follow users' instruction. Instruction-tuning dataset comes in a specific template, consisting of (1) instruction, (2) text/input and (3) answer/output as shown in Table \ref{table:template}.

\subsubsection{Multi-task Instruction Tuning}
For multi-task instruction tuning, we aligned closely with established financial benchmarks such as the pioneering FLUE benchmark \citep{shah2022flue} and the Fundamental Tasks of the FinBen \citep{xie2024finben} benchmark. We used a public dataset \texttt{Sujet-Finance-Instruct-177k}\footnote{\url{https://huggingface.co/datasets/sujet-ai/Sujet-Finance-Instruct-177k}} containing about 177,000 records that covers the following tasks in the finance domain: 
\begin{enumerate}[noitemsep,nolistsep]
    \item Sentiment Analysis: Financial Phase Bank (FPB)\citep{malo2014good}, FiQA-SA\citep{maia201818}
    \item Text classification: News Headlines Classification\citep{sinha2021impact}
    \item Named Entity Recognition (NER) \citep{alvarado2015domain}
    \item Question Answering FiQA\citep{FIQA}, ConvFinQA\citep{chen2022convfinqa}
\end{enumerate} 
For more information about this multi-task dataset refer Appendix \ref{sec:appendixA}.

The Multi-Task Instruction-tuning phase aims to train the LLMs on multiple concurrent Finance-NLP tasks, thus improving their versatility and usability on a variety of financial tasks. 


\subsubsection{Task-Specific Instruction Tuning}
In this stage, we fine-tuned the model on a single, specific task - Abstractive Text Summarization. 
Apart from solely using the training dataset (EDTSUM) provided as part of the shared task challenge, we have also identified other text summarization datasets in the Finance domain. These datasets are combined for the instruction-tuning step:
\begin{itemize}[noitemsep,nolistsep]
    \item Training set (EDTSum) (8,000 records) of financial news summarization
    \item Earnings call summarization (2,424 records)\footnote{\url{https://huggingface.co/datasets/soumakchak/earnings_call_dataset}} 
    \item Long earnings calls bullet point summarization (ECTSum)\citep{mukherjee2022ectsum}\footnote{\url{https://github.com/rajdeep345/ECTSum}}
    \item Edgar Annual Reports (10k) SEC filings summarization (1,000 records)\footnote{\url{https://huggingface.co/datasets/wyx-ucl/SUM-DATASET-BASED-EDGAR-CORPUS}}
    \item Financial news summarization (27,000 records)\footnote{\url{https://huggingface.co/datasets/kdave/Indian_Financial_News}} 
\end{itemize}


\begin{table*}[h]
    \small
  \centering
  \begin{tabular}{lllccccc}
    \hline
    \textbf{} & \textbf{Model} & \textbf{Approach} & \textbf{ROUGE-1} & \textbf{ROUGE-2} & \textbf{ROUGE-L} & \textbf{BertScore} & \textbf{BartScore}\\
    \hline
     \multirow{2}{*}{(A)} &    Llama3 8B & zero-shot & 0.2020  & 0.0921 & 0.1572  & 0.6506 & -4.0342  \\
      &    FinLlama3 & zero-shot & 0.2020  & 0.0923 & 0.1600  & 0.6701 & -4.0012  \\
        \hline
    \multirow{4}{*}{(B)} & FinMA-7B & zero-shot & 0.2021 & 0.0942 & 0.1621 & 0.5935 & -4.0941  \\
    & FinGPT(LlaMa2) & zero-shot & 0.0005 & 0.0001 & 0.0005 & 0.5128 & -4.9231  \\
    & FinLlama3\_mt & zero-shot &  0.2044 & 0.0945 & 0.1604 & 0.6517 & -3.6981  \\
    \hline
    (C) & \textbf{FinLlama3\_sum} & \textit{instruction-tuned} & \textbf{0.5210} & \textbf{0.3406} & \textbf{0.4735} & \textbf{0.9084} & \textbf{-3.4980} \\
    \hline
  \end{tabular}
  \caption{\label{table:evaluation_results} Text Summarization results tested on the EDTSum test set. (A): foundation models, (B): multi-tasks FinLLMs, (C): task-specific model. Access to InvestLM model was not available, hence not listed in the table above.}
  \vspace{-0.5em}
\end{table*}

\section{Experiments and Results}

\subsection{Training Details}

\begin{enumerate}[noitemsep,nolistsep]
    \item \textbf{Stage 1: Continual pre-training on Financial Corpus} - we further pre-train Llama3 8B\footnote{\url{https://huggingface.co/unsloth/llama-3-8b-Instruct-bnb-4bit}} on the financial corpus for 2 epochs to produce \textbf{FinLlama3}.
    \item \textbf{Stage 2: Multi-task instruction-tuning} - FinLlama3 is fine-tuned on multiple financial tasks for another 15 epochs to produce \textbf{FinLlama3\_mt}. Results and evaluation metrics for each tasks are found in Table \ref{table:multitask-result}.
    \item \textbf{Stage 3: Task-specific Instruction-tuning} - FinLlama3\_mt is then fine-tuned on text summarization task for 9 epochs. We utilize three metrics, such as ROUGE (1, 2, and L) \citep{lin2004rouge}, BERTScore \citep{zhang2019bertscore} and BARTScore \citep{yuan2021bartscore}, to evaluate the quality of the generated summaries.
\end{enumerate}


All training was done using Unsloth's FastLanguageModel\footnote{ Unsloth is a lightweight library for faster LLM fine-tuning, which is fully compatible with the Huggingface ecosystem (Hub, transformers, PEFT, TRL),\url{https://unsloth.ai/}} that uses the PEFT (Parameter Efficient Fine-Tuning) library, 4-bit quantization and QLoRA (Quantized Low-Rank Adaptation). The execution was two times faster and used 60\% less memory compared to Huggingface transformer library, significantly lower the computational requirements. As for hardware, we used one A100 GPU with 80GB memory, and with \texttt{auto-find-batch-size} set to true.

\subsection{Results and Discussion}
Apart from testing our model using the test set (EDTSUM\_test), we also ran other multi-task FinLLMs listed in Table \ref{table:FinLLM} against the same test, the results are captured in Table \ref{table:evaluation_results}. Llama3 8B is the foundation model that serves as a baseline comparison with the other FinLLMs. Among the models, only two, InvestLM \citep{yang2023investlm} and our FinLlama3\_sum were specifically instruction-tuned on the financial text summarization task. The rest of the models were evaluated for their text summarization ability via zero-shot prompting.

As expected, multi-task financial models in Group B show better performance compared to Group A's (foundation models) baseline performance. This shows that instruction-tuning a model in multiple Finance-related tasks somewhat improves the model's overall ability to generalize to unseen tasks via zero shot prompting. However, only marginal improvement is observed here. Instead, significant improvement is observed for FinLlama3\_sum that was specifically instruction-tuned for the financial text summarization task. 

The results also prove that our pipeline approach is successful as the models (FinLlama3 $\rightarrow$ FinLlama3\_mt $\rightarrow$ FinLlama\_sum) show progressive improvements in financial text summarization. While the final model, FinLlama3\_sum, performs well in text summarization, it might suffer from catastrophic forgetting on earlier fine-tuned tasks. As part of a rigorous testing, this model should be subjected to benchmark tests such as the FinBen benchmark \citep{xie2024finben}, which consist of 35 datasets across 23 financial tasks. It is important to determine if the model suffers from any catastrophic forgetting, and also to test its ability to generalize to a wider suite of financial tasks.

\section{Conclusion}
Our pipeline approach of turning a foundational model Llama3 8B into a FinLLM (LLM adapted to the Finance domain) and finally to a task-specific `expert' in Text Summarization has proven to be effective, as the final model is capable of generating summaries with a Rouge-1 score of 0.521 and is ranked 3rd in the challenge. As part of continuous improvement, we intend to explore other open-source LLMs such as Llama3 70B, and other newer models such as Gwen2 72B. 


\section{Availability}
We made our model available here: \url{https://huggingface.co/meisin123/llama3_FinLLM_textsum} and codes here: \url{https://github.com/meisin/IJCAI_FinLLMChallenge}.


\bibliography{custom}


\appendix

\section{Prompt template} \label{sec:summary_prompt}
\begin{table}[h]
\small
  \centering
  \begin{tabular*}{\columnwidth}{l}
      \hline
    \textbf{Template:}\\
    \hline
    \texttt{Instruction: [task prompt]} \textit{}    \\
    \texttt{Text: [input context]} \\
    \texttt{Answer: [output]} \\
    \hline    
    \hline
    Example for Text Summarization \\
    \hline
    \texttt{Instruction}: "You are given a text that consist of multiple \\
    sentences. Your task is to perform abstractive summarization \\on this text. Use your understanding of the content to express \\ the main 
    ideas and crucial details in a shorter, coherent, and \\ natural sounding text." \\
    \texttt{Text}: \textit{financial news article}\\
    \texttt{Answer}: \textit{abstractive summary of} \texttt{Text} \\
    \hline
  \end{tabular*}
  \caption{\label{table:template} Instruction-tuning Template for Text Summarization}
\end{table}

\section{Multi-task Instruction-Tuning}
\label{sec:appendix}

\subsection{Dataset} \label{sec:appendixA}
This dataset, \texttt{sujet-ai/Sujet-Finance-Instruct-177k}, can be accesed here: \url{https://huggingface.co/datasets/sujet-ai/Sujet-Finance-Instruct-177k}.
This dataset is made up of
\begin{itemize}[noitemsep,nolistsep]
    \item \textbf{Sentiment Analysis}: 44,209 entries. This involves analyzing financial texts to categorize sentiments as positive, negative, neutral, bearish, or bullish.
    \item \textbf{QA (Question Answering)}: 38,801 entries. Direct-answer finance questions that don't require additional context.
    \item \textbf{QA with Context}: 40,475 entries. Finance questions necessitating context for answers.
    \item \textbf{QA Conversation}: 15,613 entries. This category includes questions following a conversational history between a user and an LLM assistant.
    \item \textbf{Yes/No Question}: 20,547 entries. Questions necessitating a simple yes or no answer.
    \item \textbf{Topic Classification}: 16,990 entries. Tasks requiring the classification of financial texts into specific finance-related categories.
    \item \textbf{NER (Named Entity Recognition) Sentiment Analysis}: 962 entries. This involves conducting sentiment analysis at the entity level within texts.
\end{itemize}

\subsection{Multi-task Instruction-tuning results}
\label{sec:appendixB}

\begin{table}[h]
\small
  \centering
  \begin{tabular*}{\columnwidth}{lcc}
      \hline
    \textbf{Dataset} & \textbf{Metrics} & \textbf{Results}\\
    \hline
     Sentiment Analysis & F1 & 0.69 \\
     & Acc & 0.71 \\
     Text Classification & AvgF1 & 0.77 \\
     NER & Entity F1 & 0.57 \\
     QA & EmmAcc & 0.46 \\
    \hline    
  \end{tabular*}
  \caption{\label{table:multitask-result} Multi-task Instruction Tuning Results based on each tasks.}
\end{table}

\end{document}